# Vision-based Human Gender Recognition: A Survey

Choon Boon Ng, Yong Haur Tay, Bok Min Goi

Universiti Tunku Abdul Rahman, Kuala Lumpur, Malaysia.
{ngcb,tayyh,goibm}@utar.edu.my

**Abstract.** Gender is an important demographic attribute of people. This paper provides a survey of human gender recognition in computer vision. A review of approaches exploiting information from face and whole body (either from a still image or gait sequence) is presented. We highlight the challenges faced and survey the representative methods of these approaches. Based on the results, good performance have been achieved for datasets captured under controlled environments, but there is still much work that can be done to improve the robustness of gender recognition under real-life environments.

## 1 Introduction

Identifying demographic attributes of humans such as age, gender and ethnicity using computer vision has been given increased attention in recent years. Such attributes can play an important role in many applications such as human-computer interaction, surveillance, content-based indexing and searching, biometrics, demographic studies and targeted advertising.

Studies have shown that a human can easily differentiate between a male and female (above 95% accuracy from faces [1]). However, it is a challenging task for computer vision. Nevertheless, such attribute classification problems have not been as well studied compared to the more popular problem of individual recognition. In this paper, we survey the methods used for human gender recognition in images and videos using computer vision techniques.

In many applications, the system is non-intrusive or should not require the human subject's cooperation, physical contact or attention. Using human parts such as iris, hand or fingerprint would require some cooperation from the human and thus limit its applicability. We focus our attention on easily observable characteristics of a human. Most researchers have relied on facial analysis as a means of determining gender, while some work have been reported on using the whole body, either from a still image or using gait sequences. Also, we concentrate on approaches which make use of 2-D (rather than the more costly 3-D) data in the form of still image or videos. Audio cues such as voice are not considered in our survey as we are interested in methods using computer vision.

In general, a pattern recognition problem such as gender recognition, when tackled with a supervised learning technique, can be broken down into several steps which are object detection, preprocessing, feature extraction and classification.

In the detection phase, given an image, the human subject or face region is detected and the image is cropped. This will be followed by some preprocessing, for example to normalize against variations in scale and illumination. A widely used method for face detection is by Viola and Jones [2], which has an OpenCV implementation. The benchmark for human detection is based on using Histogram of Oriented Gradients (HOG) [3]. In the case of gait analysis, many methods use a binary silhouette of the human which is extracted using background subtraction.

In feature extraction, representative descriptors of the image are found and selection of the most discriminative features may be made. In some cases when the number of features is too high, dimension reduction can be applied. As this step is perhaps the most important to achieve high recognition accuracy, we will provide a more detailed review in later sections.

Lastly, the classifier is trained and validated with a dataset. Gender recognition is a within-object classification problem [4]. The subject is to be classified as either male or female, therefore a binary classifier is used. Examples of classifiers that have been widely used to perform gender recognition are Support Vector Machine (SVM), Adaboost, neural networks and Bayesian classifier. From our survey, SVM is the most widely used face gender classifier (usually using a non-linear kernel such as the radial basis function), followed by boosting approaches such as Adaboost. Nearest neighbor classifier and Markov models are also popular for gait-based gender classifiers.

The rest of this paper is organized as follows: In Section 2, potential applications for gender recognition is identified. Section 3, 4 and 5 review the aspects of gender recognition by face, gait and body, respectively. This is followed by concluding remarks in Section 6.

## 2  Applications

We identify several potential application areas where gender recognition would be useful. They are listed down as follows:

1. *Human-computer interaction systems.* More sophisticated human-computer interaction systems can be built if they are able to identify a human's attribute such as gender. The system can be made more human-like and respond appropriately. A simple scenario would be a robot interacting with a human; it would require the knowledge of gender to address the human appropriately (e.g. as Mr. or Miss).
2. *Surveillance systems.* In smart surveillance systems, it can assist in restricting areas to one gender only, such as in a train coach or hostel. Automated surveillance systems may also choose to pay more attention or assign a higher threat level to a specific gender.
3. *Content-based indexing and searching.* With the widespread use of consumer electronic devices such as cameras, a large amount of photos and videos are being produced. Indexing or annotating information such as the number of people in the im-age or video, their age and gender will become easier with automated systems using computer vision. On the other hand, for content-based searching such as looking for a photo of a person, identifying gender as a preprocessing step will reduce the amount of search required in the database.
4. *Biometrics.* In biometric systems using face recognition, the time for searching the face database can be cut down and separate face recognizers can be trained for each gender to improve accuracy [5].
5. *Demographic collection.* Demographic studies systems aim to collect statistics of customers, including demographic information such as gender, for example, walking into a store or looking at a billboard. A system using computer vision can be used to automate the task.
6. *Targeted advertising.* An electronic billboard system is used to present advertisements on flat panel displays. Targeted advertising is used to display advertisement relevant to the person looking at the billboard based on attributes such as gender. For example, the billboard may choose to show ads of wallets when a male is detected, or handbags in the case of female. In Japan, vending machines that use age and gender information of customer to recommend drinks have seen increased sales [6].

## 3 Gender Recognition by Face

Facial images are probably the most common biometric characteristic used by humans to make a personal recognition [7]. The face region, which may include external features such as the hair and neck region, is used to make gender identification.

### 3.1 Challenges

The image of a person's face exhibits many variations which may affect the ability of a computer vision system to recognize the gender. We can categorize these variations as being caused by the human or the image capture process.

Human factors are due to the characteristics of a person, such as age, ethnicity and facial expressions (neutral, smiling, closed eyes etc.), and the accessories being worn (such as eye glasses and hat). Factors due to the image capture process are the person's head pose, lighting or illumination, and image quality (blurring, noise, low resolution). Head pose refers to the orientation of the head relative to the view of the image capturing device. The human head is limited to three degrees of freedom, as described by the pitch, roll and yaw angles [8].

The impact of age and ethnicity on the accuracy of gender classification has been observed. Benabdelkader and Griffin [9], after testing their classifier with a set of 12,964 face images, found that a disproportionately large number of elderly females and young males were misclassified. In empirical studies by Guo et al. [10] using several classification method on a large face database, it was found that gender classification accuracy was significantly affected by age, with adult faces having higher accuracies than young or senior faces. In [11], when a generic gender classifier trained for all ethnicities was tested on a specific ethnicity, the result was not as good as a classifier trained specifically for that ethnicity.

### 3.2 Preprocessing

After the face is segmented from the image, some preprocessing may be applied. It helps to reduce the sensitivity of the classifier to variations such as illumination, pose and detection inaccuracies. Graf and Wichmann [12] pointed that cues such as brightness and size will be learnt by the classifier such as SVM to produce artificially better performance.

Preprocessing that may be applied to the face image include:

− Normalize for contrast and brightness (e.g. using histogram equalization)
− Removal of external features such as hair and neck region
− Geometric alignment (either manually or using automatic methods)
− Downsizing to reduce the number of pixels

For efficiency, it is preferable that the face image does not undergo alignment as it requires significant time [13]. In a study by Mäkinen and Raisomo [14], it was found that automatic alignment methods did not increase gender classification rate while manual alignment increased the classification rate a little. They concluded that automatic alignment methods need to be improved. Also, alignment is best done before downsizing. If alignment is not done, deliberately adding misaligned faces to the training data seems to help make the classifier robust to face misalignments [15].

### 3.3 Facial feature extraction

We broadly categorize feature extraction methods for face gender classification into *geometric-based* and *appearance-based* methods, following [9][16]. The former is based on measurements of facial landmarks. Geometric relationships between these points are maintained but other useful information may be thrown away [9] and the process of extracting the point locations need to be accurate [17]. Appearance-based methods are based on some operation or transformation performed on the pixels of an image. This can be done at the global (holistic) or local level. At the local level, the face may be divided into defined regions such as eyes, nose and mouth or regularly spaced windows. In appearance-based methods, the geometric relationships are naturally maintained [9], which is advantageous when the gender discriminative features are not exactly known. However, they are sensitive to variations in appearance (due to view, illumination, expression, etc.) [9] and the large number of features [17]. All methods mentioned in the following, other than those under *fiducial distances*, can be categorized as appearance-based.

**Fiducial distances**. Important points on the face that mark features of the face, such as nose, mouth, hair, ears and eyes are called facial landmarks or fiducial points. Fiducial distances are the distances between these points. Psychophysical studies using human subject established the importance of these distances in discriminating gender. Brunelli and Poggio [18] used 18 point-to-point distances to train a hyper basis function network classifier. Fellous [19] selected 40 points to calculate 22 normalized vertical and horizontal fiducial distances. These points were extracted manually from images of frontal faces by a human operator. From these distances, five dimensions were derived using discriminant analysis and used to classify gender. Mozzafari et al. [20] used the aspect ratio of face ellipse fitting and rms distance between the ellipse and face contour as geometric features to complement their appearance-based method.

**Pixel intensity values.** Pixel intensity values are used directly as input to train a classifier such as neural network (e.g. in the early works of [21] [22] [23]) or support vector machine (SVM). As a preprocessing step, the images (after cropping the head), are usually normalized to compensate for geometric and lighting variations, and finally down-sampled to lower sizes. Gutta et al. [24] used a mixture of experts consisting of ensembles of radial basis functions (RBFs) combined with inductive decision trees for classification of 64 x 72 pixel face images. Moghaddam and Yang [25] used 21 x 12 pixel images to train an SVM classifier. Gaussian RBF kernel was found to give the best performance for the SVM. Their results were considered as state-of-the-art for some time. Pyramidal neural network architecture [26] and shunting inhibitory convolutional neural network [27] have also been used as classifier. These work used down-sampled frontal face images.

Baluja and Rowley [28] proposed a fast method that matched the accuracy of SVM classifier. Simple pixel comparison operations were used as features for weak classifiers which were combined using AdaBoost to achieve performance 50 times faster. The face images were normalized to 20 x 20.

Using pixel intensity values result in a large number of features which increases proportionately to the size of the image. Dimension reduction methods such as Principal Component Analysis (PCA) obtain a representation of an image in reduced dimension space. Early studies on gender recognition [22][21] used PCA. Genetic Algorithm (GA) was used by Sun et al. [29] to remove eigenvectors that did not seem to encode gender information. Castrillón et al. [30] used the PCA representation to train a SVM and used majority vote for temporal fusion in video sequences obtained from webcam recording. A study by Buchala et al. [31] found that different components of PCA encode different properties of the face such as gender, ethnicity and age. Bui et al. [32] combined the vectors obtained from pixels of the whole face image, local face regions and the gradient image. The vectors are combined to form a single vector on which PCA was then applied.

Two-dimensional PCA (2DPCA) [33] has also been used for dimension reduction. Lu and Shi [34] applied 2DPCA on three facial regions to obtain features. SVM was used on each region and the classification result was based on consensus decision.

Another dimension reduction method, Independent Component Analysis (ICA), was studied by Jain et al. [35]. Curvilinear Component Analysis (CCA) was proposed by Demartines and Herault [36] to reduce the dimensionality of nonlinear data. Bu-chala et al.[37] showed that CCA reduced the dimension of face image data more effectively compared to PCA with comparable gender classification rate.

**Rectangle features.** Viola and Jones[38] introduced rectangle features for rapid face detection. Figure 1 shows the example of these rectangles, which are also known as Haar-like features. The sum of the pixels which lie within the white rectangles are subtracted from the sum of pixels in the grey rectangles to obtain the value of a rectangle feature. Integral image representation can be used to compute the rectangle features rapidly. Adaboost is used to select the features and produce a perceptron classifier. Shakhnarovich et al. [13] used these features for fast gender and ethnicity classification of videos in real-time. Xu et al. [39] combined it with fiducial distances.

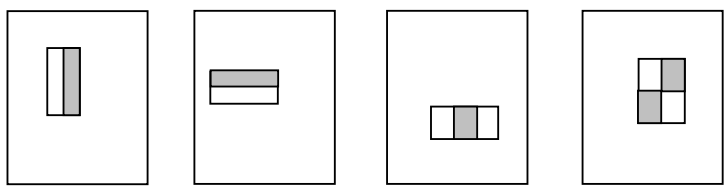

**Fig. 1.** Rectangle features [38]

**Local binary patterns.** Ojala et al. [40] introduced local binary patterns (LBP) for grayscale and rotation invariant texture classification. Each pixel in an image is labeled by applying the LBP operator, which thresholds the pixel's local neighborhood at its grayscale value into a binary pattern. The local neighborhood is a circular symmetric set of any radius and number of pixels. A subset of the patterns, called "uniform patterns", is defined to be patterns with at most 2 bitwise transitions (0/1 or 1/0). They detect microstructures such as edge, corners and spot. The histogram of these patterns is then used as a feature to describe texture. Lian and Lu [41] used LBP with SVM for multi-view gender classification while Yang and Ai [42] applied it for classifying age, gender and ethnicity. Alexandre et al. [43] combined LBP with intensity and shape feature (histogram of edge directions) in a multi-scale fusion approach, while Ylioinas et al. [44] combined it with contrast information. Shan [45] used Adaboost to learn discriminative LBP histogram bins.

Other variants inspired by LBP have been proposed, such as Local Gabor Binary Mapping Pattern [46][47][48], centralized Gabor gradient binary pattern [49], Local Directional Pattern [50] and Interlaced Derivative Pattern [51].

**Scale Invariant Feature Transform (SIFT).** SIFT features are invariant to image scaling, translation and rotation, and partially invariant to illumination changes and affine projection [52]. Using these descriptors, objects can be reliably recognized even from different views or under occlusion. The advantage of using invariant features such as SIFT is that the preprocessing stage, including accurate face alignment, is not required [53]. Demirkus et al. [54] exploited these characteristics, using a Markovian model to classify face gender from unconstrained video in natural scenes. Wang et al. [55] extracted SIFT descriptors at regular image grid points and combined it with global shape contexts of the face, adopting Adaboost for classification. In another work, they combined the SIFT descriptors with Gabor features [56].

**Gabor wavelets.** Research in neurophysiology has shown that Gabor filters fit the spatial response profile of certain neurons in the visual cortex of the mammalian brain. Lee [57] derived a family of 2-D Gabor wavelets for image representation. A Gabor wavelet is defined by frequency, orientation and scale. Wiskott et al. [58] used Gabor wavelets to label the nodes of an elastic graph which was used to represent the face. Lian et al. [59] following the method by Hosoi et al [60], used Gabor wavelets from different facial points located using retina sampling. Leng and Wang [61] extracted Gabor wavelets of five different scales and eight orientations from each pixel of the image as features, which were then selected using Adaboost. Scalzo et al. [62] extracted a large set of features using Gabor and Laplace filters which are used in a feature fusion framework of which the structure was determined by genetic algorithm.

Gabor filters have also been used to obtain the simple cell units in biologically inspired features (BIF) which was proposed by Riesenhuber and Poggio [63] for object recognition and later extended by Meyers and Wolf [64] for face processing. This model contains simple (S) and complex (C) cell units arranged in hierarchical layers of S1, C1, S2 and C2, with an S2FF layer for face processing. Guo et al. [10] found that, for face gender recognition, the C2 and S2 layers degraded performance.

**Others.** Other facial representations that have been used include a generic patch-based representation [4], regression function [65], multiscale filter banks in the BIF model [66], Ranking Labels [67], subspace of random crops [68], Discrete Cosine Transform [69][70], wavelets of Radon transform [71] and intensity statistics such as mean, variance, skew and kurtosis [15].

For work using non-visible light images, experiment results of [72] indicate the possibility of performing gender classification using near infrared images although the performance was slightly inferior to visible spectrum images.

**External cues.** Features external to the face region such as hair, neck region and clothes are also cues used by humans to identify gender. Ueki and Kobayashi [73] integrated color and edge features from the face and neck regions, since the former contains factors such as neck size, jawline and clothing type. Li et al. [16] included features from hair and upper body clothing. Gallagher and Chen [74] used social context information based on position of a person's face in a group of people to help infer gender in photographs.

### 3.4 Face Datasets

Table 1 summarizes several publicly available datasets that have been used for evaluating gender recognition. These datasets tend to be those collected for the use in face recognition or detection evaluation. Some researchers take only a subset of the datasets (excluding unsuitable images), or in order to obtain a large amount of images, combine several datasets, including using their own collection (for example, collected from the web). One common practice is to use a face detector to obtain face crops. However, this may cause the data to be biased, for example if the detector successfully detects only frontal and near-frontal faces. None of these public datasets were designed specifically for gender recognition evaluation. As for private datasets, examples are WIT-DB [73], BUAA-IRIP [61], UCL [75], YGA [10] and BCMI[16].

**Table 1.** Public Face Datasets

| Dataset | No. of images | No. of unique Individuals | Controlled variations |
|---|---|---|---|
| AR [76] | >4000 | 126 (70m,56f)[#] | X,L,O |
| XM2VTS [77] | 5900 | 295 | P,L |
| FERET [78] | 14126 | 1199 | P,L,X |
| BioID [79] | 1521 | 23 | L, face size, background |
| CMU-PIE [80] | 41368 | 68[#] | P,L,X |
| FRGC [81] | 50000 | 688 | L,X, background |
| UND Biometrics-B [82] | 33287 | 487 | L,X |
| MORPH-2 [83] | 55285 | 13660 (46767m, 8518f)[#] | age |
| LFW [84] | 13233 | 5749 (10256m, 2977f) | Uncontrolled |
| CAS-PEAL-R1 [85] | 30900 | 1040 (595m, 445f)[#] | P,X,L,O |
| Images of Groups [74] | 5080 | < 28231[#] | Uncontrolled |

*Notes on the table*:-
Under *No. of unique individuals*, the breakdown of male and female faces is given in brackets, where known; for example, 500m, 500f refers to 500 male and female faces each.
[#] indicates gender is labeled.
The *controlled variations* are indicated as follows:
P – pose, view     L – lighting, illumination     X – expression     O – occlusion

FERET is a widely used dataset for evaluation of face recognition algorithms, and has also been used by many researchers for face gender recognition. It contains 14,126 images of 1199 individuals [78]. The faces have a variety pose, and some variation in illumination and expression. Baluja et al. [28] noted that good results were achieved with this database because the images are noise-free, have consistent lighting, and without background clutter. Also, earlier researchers may have made the mistake of including the same person in both the training and test set. The gender information is not labeled, although for a subset of images (with 212 males and 199 females), it has been made available by Mäkinen and Raisomo [14].

The CAS-PEAL face database is a large scale Chinese face database which contains 99,594 images of 1040 individuals (595 males and 445 females) with varying pose, expression, accessories and lighting [85]. A subset of the database CAS-

PEAL-R1 which contains 30,900 images is available for research purpose by request. The images were taken in a controlled environment with lighting and cameras placed at various angles. Various accessories (glasses and hats) were worn by the subjects and they were also asked to make various expressions. Different backgrounds were also used to capture the effect of change in white balance of the camera. Gender information is contained in the image file name.

LFW (Labeled Faces in the Wild) was compiled to aid the study of unconstrained face recognition. The dataset contains faces that show a large range of variation typically encountered in everyday life, exhibiting natural variability in factors such as pose, lighting, race, accessories, occlusions, and background [84]. The number of males outnumbers females, with some individual having appearing more than once.

In many datasets, the images are not annotated with gender information. Therefore researchers had to manually label the ground truth using visual inspection, either by themselves or with the help of others.

As a conclusion, no large, publicly available dataset specifically designed for the problem of face gender recognition has been established. Recently, the MORPH-2 [83] and LFW datasets has been proposed as the standard for controlled and uncontrolled face gender classification, respectively [86]. Dago-casas et al. [87] recommend Gallagher's Images of Groups dataset over LFW for uncontrolled conditions, as the former is more gender-balanced.

### 3.5 Evaluation and results

A list of representative works on face gender recognition is compiled in Table 2. The table compares the features and classifier used, training and test dataset used and average total classification rate, as reported by the authors. The dataset characteristic in terms of the controlled variety of images are also indicated.

The average classification rate (also referred to as classification accuracy or recognition rate) obtained from the results of cross-validation is usually reported. The classification rate is the ratio of correctly classified test samples to the total number of test samples. Five-fold cross validation is often used. Some researchers may test their method on different databases for generalization ability (i.e. train with dataset A, test with dataset B.) It is good practice to ensure faces of the same individual do not appear on both the training and test set. This is to prevent the classifier from recognizing individual faces rather than gender [28]. Some researchers also ensure the same number (or ratio) of male and female faces are kept in the training and test sets.

Because of the different datasets and parameters used for evaluation, a straight comparison between the methods is difficult. Some researchers use frontal face images only, while other may include non-frontal faces (variation in pose or view). The variation (e.g. age, expression, illumination, etc.) that are present in the datasets also differ. We have provided a review of some of the datasets in the previous section. It is noted that the FERET dataset is the most often used. Even then, the subset of images used varies between researchers.

**Table 2.** Face gender recognition

| First Author, Year | Feature extraction | Classifier | Training data | Test data | Ave. Acc.% | Dataset variety |
|---|---|---|---|---|---|---|
| Gutta, 2000 [24] | Pixel values | RBF + Decision tree | FERET- fa, fb 1906m 1100f | k-fold CV | 96 | F,E |
| Moghaddam, 2002 [25] | Pixel values | SVM-RBF | FERET 1044m 711f | 5-CV | 96.62 | F |
| Shakhnarovich, 2002 [13] | Haar-like | Adaboost | Web images | 5-CV Video seqs. | 79 90 | P (<30˚), A,E,L |
| Sun, 2002 [29] | PCA with GA | SVM | UNR 300 m 300f | 3-CV | 95.3 | E,X,L,S |
| Castrillon, 2003 [30] | PCA | SVM+ temporal fusion | Video frames 798m 231f | 8123m 1755f | 98.57 | U,F |
| Buchala, 2005 [88] | PCA | SVM –RBF | Mix (FERET, AR, BioID) 200m 200f | 5-CV | 92.25 | F |
| Jain, 2005 [35] | ICA | SVM | FERET 100m 100f | FERET 150m 150f | 95.67 | F,S |
| Baluja, 2006 [28] | Pixel comp. | Adaboost | FERET-fa,fb 1495m 914f | 5-CV | 94.3 | F,S |
| Lapedriza, 2006 [66] | BIF multi scale filt. | Jointboost | FRGC 3440t FRGC 1886t | 10-CV 10-CV | 96.77 91.72 | Uniform background Cluttered background |
| Lian, 2006 [41] | LBP histogram | SVM-polynomial | CAS-PEAL 1800m 1800f | CAS-PEAL 10784t | 94.08 | P (up to 30˚ yaw & pitch) |
| Fok, 2006 [26] | Pixel values | Convolutional neural net. | FERET - fa 1152m 610f | 5-CV | 97.2 | F |
| Yang, 2007 [42] | LBP histogram | Real Adaboost | Chinese shots 4696m, 3737f | 5-CV FERET 3540t PIE 696t | 96.32 93.3 91.1 | U(X,O) F,X,O F,X,O |
| Makinen, 2008 [5] | Various - pixels, LBPH, Haar-like | Various (ANN, SVM, Adaboost) in combination | FERET 304m 304f web images 1523m 1523f | FERET 76m 76f web images 381m 381f | 92.86 83.14 | F,S F |
| Leng, 2008 [61] | Gabor | Fuzzy SVM | FERET 160m 140f CAS-PEAL 400m 400f BUAA-IRIP 150m 150f | 5-CV 5-CV 5-CV | 98 89 93 | F F F |
| Xu, 2008 [39] | Haar-like, fiducial distances | SVM-RBF | Mix ( FERET, AR,Web) 500m 500f | 5-CV | 92.38 | F,E,A,L,S |
| Xia, 2008 [46] | LGBMP hist. | SVM-RBF | CAS-PEAL 1800m 1800f | CAS-PEAL 10784t | 94.96 | P (up to 30˚ yaw & pitch) |
| Scalzo, 2008 [62] | Gabor & Laplace | kernel spectral regression | UNR 200 m 200f | 3-CV | 96.2 | E,X,L,S |
| Zafeiriou, 2008 [89] | Pixel values | SVM variant | XM2VTS 1256m 1104f | 5-CV | 97.14 | S |
| Aghajanian, 2009 [4] | Patch-based | Bayesian | Web images 16km 16kf | Web images 500m 500f | 89 | U |
| Li, 2009 [70] | DCT | Spatial GMM | YGA 6096t | YGA 1524t | 92.5 | F,A,S |
| Lu, 2009 [34] | 2D PCA | SVM-RBF | FERET 400m 400f CAS PEAL 300m, 300f | 5-CV CAS-PEAL 1800t | 94.85 95.33 | F F,X |
| Demirkus, 2010 [54] | SIFT | Bayesian | FERET 1780m 1780f | Video seqs. (15m 15f) | 90 | U (P,X,O,L) |
| Wang, 2010 [56] | SIFT, Gabor | Adaboost | Mix (FERET, CAS-PEAL, Yale, I2R) 4659t | 10-CV | ~97 | F,X,L,O |
| Lee, 2010 [65] | regression function | SVM | FERET-fa 1158m, 615f Web images 3000 t | 5-CV Web images 3000t | 98.8 88.1 | F A,E |
| Alexandre, 2010 [43] | Intensity, hist. of edge dir., LBP | SVM-linear | FERET 152m 152f UND set B 130m 130f | FERET 60m 47f UND set B 171m 56f | 99.07 91.19 | F,S F,S |
| Li, 2011 [16] | LBP, (+ hair & clothing features) | SVM | FERET 227m 227f BCMI 821m 821f | FERET 114m 114f BCMI 274m 274f | 95.8 95.3 | F F |
| Wu, 2011 [47] | LGBP | SVM-RBF | CAS-PEAL 2142m 2142f | CAS-PEAL 2023m 996f | ~91-97 per set | P (up to 67° yaw), S |
| Zheng, 2011 [48] | LGBP-LDA | SVMAC | CAS-PEAL 2706m 2706f (of 9 sets) FERET 282m 282f BCMI 361m 361f | CAS-PEAL 2175m 1164f FERET 307m 121f BCMI 168m 155f | ≥ 99.8 per set 99.1 99.7 | P (up to 30° yaw & pitch), S F F |
| Shan, 2012 [45] | LBP hist. bins | SVM-RBF | LFW 4500m 2943f | 5-CV | 94.81 | F,U,S |

*Notes on the table*:-
*Training data* and *testing data* gives information on the dataset from which the images were taken for training and testing the classifier, respectively. The breakdown of male and female faces is also given; for example, 500m 500f refers to 500 male and female faces each. Where the breakdown could not be determined or was not given, the total faces used are given (e.g. 1000t).
When the classification rate or accuracy is based on cross-validation result, this is indicated in the *testing data* field; for example, 5-CV refers to five-fold cross validation, and the average rate from validation results are given in the *Ave. Acc.* field. If classification rate for a separate or different test set is given, this is used and the dataset is indicated.
Under *dataset variety*, the variations controlled or the variety available in the dataset, as mentioned by the authors, are indicated as follows:
F – frontal only     A – age     E– ethnicity     P – pose, view     L – lighting, illumination
X – expression     O – occlusion     U – uncontrolled
S – indicates the same individual does not appear on both training and test set

For face images from FERET dataset, the best result is obtained by Zheng et al. [48] and Alexandre et al. [43], with a classification rate of 99.1%. However, only frontal faces from the dataset were used. Zheng et al. [48] achieved near 100% for pose variations up to 30° yaw and pitch on the CAS-PEAL dataset. However, separate classifiers had to be trained for each pose. For images taken in uncontrolled environments, Shan [45] obtained 94.8% on the LFW dataset which contains frontal and near frontal faces.

In order to standardize the evaluation of facial analysis techniques, including the gender classification problem, an international collaborative effort named BeFIT [90] has promoted benchmarking activities by proposing standard evaluation protocols and datasets. The MORPH-II and LFW datasets were proposed for constrained and unconstrained face gender classification, respectively. Five-fold cross validation should be used, with images of individual subjects only in one fold at a time. Distribution of age, gender and ethnicity in the folds should be similar as the whole dataset. More robust evaluation metrics (in addition to the traditionally-used metric mentioned above) were recommended – TPR (*true positive rate*), TNR (*true negative rate*) and ACR (*average correct rate*, defined as the average of TPR and TNR). To deal with datasets having imbalanced gender, AUC (*area under the receiver operator characteristic curve*) should be used [86]. It will be interesting to compare the state-of-the-art methods based on this recently established protocol.

### 3.6     3-D approaches

In this section, we briefly review 3-D approaches. Some studies have shown that gender information is also contained in the depth profile of the face [12] [91] and that gender classification is more effective with three dimensional head structure than with image intensity information [92]. The shape information in 3-D face data (in the form of points or meshes) from head scans is exploited by several researchers. Han et al. [93] calculated the ratio of the surface area and volume of prominent facial features (eyes, nose, mouth, cheek) in comparison to the whole face. Toderici et al. [94] used Haar wavelet coefficients derived from the geometry- image representation of a fitted deformable model. Lu et al. [95] combined registered range data (3-D points from head scans) with intensity images, using SVM as classifier, to show better results compared to each individual modality.

Appearance-based methods use 2-D images projected from the 3-D data obtained from scanners. Tariq et al. [96] and Yang et al. [97] used the contour of the face profile. Shen et al. [98] found that LBP was more effective than the raw pixels of the 3-D images. Hu et al. [99] extracted 15 facial landmarks using 3-D information and then obtained five facial regions to train SVM classifiers for each region.

These approaches using 3-D data have the disadvantage of requiring expensive scanners and high computational complexity [100]. A 2.5D representation based on facial surface normals, called facial needle-maps [101], can be retrieved from 2D images and contain 3-D facial shape information from a fixed view point [100]. The method requires a statistical face model constructed from laser scan data.

## 4     Gender Recognition by Gait

Gait is defined to be the coordinated, cyclic combination of movements that result in human locomotion [102]. This would include walking, running, jogging and climbing stairs. However, in computer vision research, it is usually restricted to walking. The main advantages of using gait as a biometric are that it is non-obtrusive and can be captured at a distance [103], in public places, without requiring cooperation or even awareness of the subject [102]. The use of gait as a biometric is considered relatively young history when compared to methods that use voice, finger prints or faces [102].

Classifying gender based on gait can be useful in some situations such as when the face is not clear, not visible or heavily occluded.

### 4.1 Challenges

Many factors affect the gait of a person, such as load, footwear, walking surface, injury, fatigue, drunkenness, mood, and change with time. Video-based analysis of gait would also need to contend with the person's clothing, camera view, speed of walking and background clutter. For example, Hu et al [104] obtained lower gender classification rates when the subject is carrying a bag or wearing overcoat. Makihara [105] found that better results could be achieved if the age group was restricted to young adults. The problem of view dependence has also been studied by many researchers [105][106][107][108][109][110][111].

In a video sequence of a person walking, the gait cycle can be referred to as the time interval between two consecutive left/right mid-stances [103]. Thus, a sequence could contain one or more gait cycles. Gait is thus a dynamic biometric that contains additional temporal and frequency information.

### 4.2 Feature extraction

Early work on gait analysis used point lights attached to the body's joints. It was found that based on the motion of the point lights during walking, identity and gender of a person could be identified.[112][113][114] The reader can refer to [115] for a survey on these early works. In this section, we review the representation and features that have been used for gait-based gender recognition.

Human gait representation can be divided into *appearance-based* (*model-free*) or *model-based* [116][117]. Appearance-based approaches have lower computational cost while model-based approaches suffer from difficulty in extracting the features robustly as they rely on accurate estimation of joints [117][118] and require high quality gait sequences [103] where the body parts need to be tracked in each frame. Moreover, model-based approaches ignore width information of the human body [118]. However, they are view and scale invariant [103].

**Model-based.** Yoo et al., [119] obtained 2D stick figures from the body contour, guided by anatomical knowledge. The sequence of stick figures from one gait cycle was taken as a gait signature (Figure 2). Temporal, spatial and kinematic parameters, and moments were considered as features and selected using a statistical distance measure.

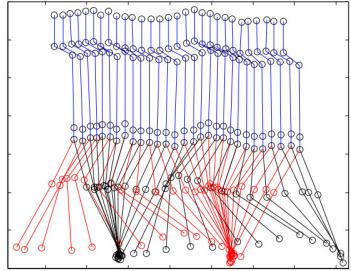

**Fig. 2.** Gait signature [119]

**Appearance-based.** In many methods, a silhouette of the walking human is obtained first from the images in a gait sequence. Lee and Grimson [120] divided each human silhouette into 7 regions and fitted ellipses into each region (Figure 3). The mean and standard deviation of the ellipse centroid, major axis orientation and aspect ratio of major and minor axis, together with the centroid height of the whole silhouette, was taken across time to form the gait average appearance features. The features were then selected using ANOVA. The advantage of the feature is robustness to silhouette noise. However, it will be affected by viewpoint, clothing and gait changes [120]. Felez et al. [121] improvised by using a different regionalization of 8 parts to obtain more realistic ellipses and meaningful feature space. However, such regional features are vulnerable to deformed silhouettes caused by frequent occlusion [118]. Thus, equal partitions formed by 2x2 and 4x4 grids were used by Hu et al. [118]. The ellipse fit parameters were fused with the stance indexes as spatial and temporal features respectively to train a mixed conditional random field (MRCF).

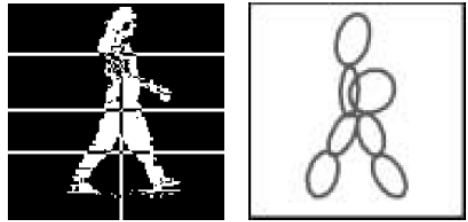

**Fig. 3.** Partitioned silhouette and the fitted ellipses [120]

Zhang and Wang [110] used *frieze patterns* of Liu et al. [122] to study multi-view gender classification, in which they analyzed and compared the class separability of these features from different view angles Fisher linear discriminant (FLD) analysis. A *frieze* pattern is a two-dimensional pattern that repeats along one dimension (Figure 4). The gait representation is generated by projecting the silhouette along its columns and rows, then stacking these 1-D projections over time [122]. These patterns enable viewpoint estimation and can be used in model-based analysis for locating body parts in video frames.

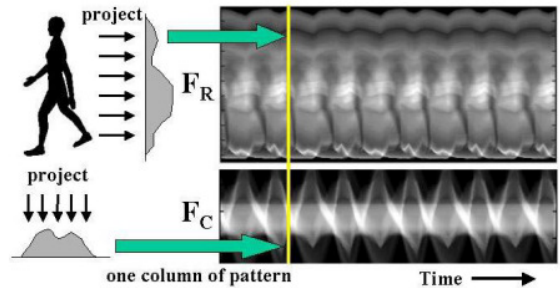

**Fig. 4.** Frieze patterns [122]

Shan et al. [123] showed that the *Gait Energy Image* (GEI) by Han & Bhanu [124] was an effective representation for gender recognition. They fused gait with face features using canonical correlation analysis (CCA) to improve performance. A GEI [124] represents human motion in a single image while preserving temporal information. It is obtained by averaging the silhouette images in one or more gait cycles to produce a grayscale image (Figure 5), thus saving on storage and computational cost. GEIs are also robust to silhouette noise in individual frames. Liu and Sarkar [125] proposed a similar representation, averaged on one gait cycle, which they called the average gait image (AGI).

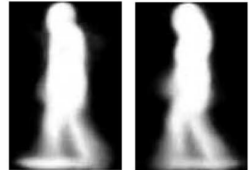

**Fig. 5.** Two examples of GEI [124]

Yu et al. [117] divided the GEI into 5 different components, with each given a weight based on the results from psychophysical experiments. Li et al. [126] partitioned the AGI into 7 components corresponding to body parts, while Chen et al. [109] used 8 components based on their consideration of walking patterns. Chang et al.[107] obtained the GEI by estimating from a whole gait sequence, thus eliminating the need to detect the gait cycle frequency. Hu and Wang [127] proposed a novel gait pattern called Gait Principal Component Image (GPCI) which was obtained using PCA. Lu and Tan [106] obtained the difference GEI from different views and introduced uncorrelated discriminant simplex analysis (USDA) to project the GEIs into lower dimensional feature subspace to increase view-invariant gender discrimination.

Signal transformation methods such as Radon, Fourier and wavelet transforms have also been used. Chen et al. [128] applied Radon transform on the human silhouettes in a gait cycle and then used Relevant Component Analysis (RCA) for feature transformation. Oskuie and Faez [129] applied Radon Transform on the Mean

Gait Energy Image of Chen et al. [130] and then extracted the Zernike moments. Makihara et al. [105] used frequency-domain features obtained from the silhouette using Discrete Fourier Transform (DFT) and reduced the dimension using Singular Value Decomposition. Handri et al. [131] applied wavelet decomposition using Daubechies function on the silhouette contour width and found it performed better than Fourier Transform.

Some methods work directly on the image of the human subject instead of extracting the silhouette. Chang and Wu [108] used DCT coefficients as texture features to train a classifier based on embedded hidden Markov models. Hu et al. [104] applied Gabor filter banks of 3 different scales and 6 orientations on the image to extract the features. Maximization of Mutual Information (MMI) was used to learn the discriminative low dimensional representation. The Gabor-MMI feature vectors were used to train two Gaussian Mixture Model-Hidden Markov Models (GMM-HMMs) to perform classification.

There have also been work based on fusion of face and gait features. [123][132]. Both of these work use frontal face and side view of the gait to extract features, which implies that 2 cameras would be needed in a real-world implementation.

### 4.3 Gait datasets

Datasets designed for evaluation of gait recognition were used for gender recognition. Table 3 gives a summary of these datasets.

**Table 3.** Gait datasets

| Dataset | No. of sequences | No. of unique individuals | Controlled variations | Remarks |
|---|---|---|---|---|
| SOTON Large Gait [133] | >600 | >100 | View (front, side), Scene (outdoor, indoor, treadmill) | only 15 females [117] |
| Human ID [134] | 1870 | 122 (85 males, 37 females) | View, shoe, walking surface, carry briefcase, elapsed time | |
| CASIA Gait Set B [135] | 13640 | 124 (93 males, 31 females) | View (11 azimuths), Walking status (normal, carry bag, wear overcoat) | Gender labeled |
| BUAA-IRIP Gait [132] | 4800 | 60 (32 males, 28 females) | View (7 azimuths) | private |
| OU-ISIR [105] | >4200 | 168 (88 males, 80 females) | View (2 heights, 12 azimuths and overhead), Age | only a portion in public |

The Human ID dataset from University of South Florida contains 1870 sequences of 122 subjects. Each subject was asked to walk multiple circuits around an ellipse in outdoors, with the last circuit taken as the dataset. However, the set is imbalanced in gender, with more 85 males and only 37 females.

The Institute of Automation, Chinese Academy of Sciences (CASIA) produced the CASIA Gait Database for gait recognition research, consisting of 3 sets- Dataset A, Dataset B and Dataset C. Dataset A is a small database with 20 persons only, while

Dataset C are images taken using thermal infrared camera. Dataset B has 124 persons of mostly Asians (except 1 European), and more males and females (93 vs. 31).

The BUAA-IRIP gait dataset was introduced by Zhang et al. [110] to study multi view gender recognition using gait. It contains an almost equal number of male and females, but the dataset is currently not publicly available. Recently, Osaka University introduced the OU-ISIR dataset which has 168 subjects of almost the same number of males and females. The subjects cover a large range of age groups, from 4 to 75 years old. Currently, only a portion of the dataset is available for download, in the form of silhouette sequences.

As a summary, compared to face datasets, gait datasets are currently smaller in the number of subjects, perhaps due to acquisition costs. There is a need for publicly available datasets with a larger number of subjects, balanced in gender, and also with gait sequences captured in uncontrolled environments.

### 4.4 Evaluation and results

Table 4 shows a list of works on gender recognition based on gait. The table compares the feature extraction methods, classifier, training and test dataset, and average total classification rate, as reported by the authors. The dataset characteristic in terms of the controlled variety of sequences are also indicated. Generally, the average correct classification rate obtained from the results of cross-validation is reported.

**Table 4.** Gait-based gender recognition

| First Author, Year | Feature Extraction | Classifier | Training data | Test data | Ave. Acc.% | Dataset variety |
|---|---|---|---|---|---|---|
| Lee, 2002 [120] | Ellipse fittings | SVM | Private 14m 10f (194) *evenly split | | 84.5% | N |
| Yoo, 2005 [119] | 2D stick figures | SVM-polynomial | SOTON 84m 16f | 10-CV | 96% | N |
| Huang, 2007 [111] | ellipse fittings | SVM | CASIA B 25m 25f (300) | CASIA B 5m 5f(60) | 89.5% | M (0°, 90°, 180°) |
| Shan, 2008 [123] | GEI + PCA +LDA | Nearest neighbour | CASIA B 88m 31f (2380) | 5-CV | 94.5 | N |
| Chen, 2009 [128] | Radon transform silhouette + Relevant component analysis | Mahalanobis distance | IRIP 32m 28f (300) | LOO-CV | 95.7 | N |
| Chen, 2009 [109] | AGI | Euclidean distance | IRIP 32m 28f (300 per angle) | LOO-CV | 93.3 (~73-92, per view) | M(0°-180) |
| Yu, 2009 [117] | GEI | SVM | CASIA B 31m 31f (372) | 31-CV | 95.97 | N |
| Chang, 2009 [107] | GEI+ PCA+LDA | Fisher boost | CASIA B 93m 31f (8856) | 124-CV Videos 2m 2f(32) | 96.79 84.38 | M(0°-180°) M(U) |
| Hu, 2009 [127] | GPCI | k-NN | IRIP 32m 28f (300) | LOO-CV | 92.33 | N |
| Chang, 2010 [108] | DCT | EHMM | CASIA B 25m 25f | 5-CV | 94 | M(0°-180°) |
| Lu, 2010 [136] | GEI + UDSA | Nearest neighbour | CASIA B 31m 31f (4092) | LOO-CV | 83-93 (per view) | M(0°-180°) |

| Felez, 2010 [121] | Ellipse fittings | SVM-linear | CASIA B 93m 31f (744) | 10-CV | 94.7 | N |
| --- | --- | --- | --- | --- | --- | --- |
| Hu, 2010 [104] | Gabor + MMI | GMM-HMM | CASIA B 31m 31f (372) | 31-CV | 96.77 | N |
| Hu, 2011 [118] | ellipse fittings & stance indexes | MRCF | CASIA B 31m 31f (372) | 31-CV | 98.39 | N |
| | | | IRIP 32m 28f (300) | LOO-CV | 98.33 | N |
| Handri, 2011 [131] | Wavelet trx. of silhoutte contour width + Modest adaboost | k-NN | Private 29m 14f (>172) | LOO-CV | 94.3 | N, A |
| Makihara, 2011 [105] | DFT of silhouette + SVD | k-NN | OU-ISIR 20m 20f | 20-CV | ~70-80 (per view) | M (0˚-360˚, overhead) |
| Oskuie, 2011 [129] | RTMGEI + Zernike momts. | SVM | CASIA B 93m 31f | | 98.5 | N |
| | | | CASIA B 93m 31f | | 98.94 | N, W, C |

*Notes on the table*:-
Refer to notes for Table 2 for information regarding *Training data*, *Test data* and the *Ave. Acc.* fields.
Under *Test data*, the figure in the bracket is the total number of sequences used.
LOO-CV refers to leave-one-out cross validation.
Under *dataset variety*, the variations controlled are indicated as follows:
   N – side view only       M– multi-view (the range of angles are also given)
   A – various age          W – wearing overcoat      C– carrying bag

For the CASIA gait dataset, Hu et al. [118] reported state of the art performance of 98.39% using side view sequences only. Oskuie and Faez [129] achieved slightly higher result of 98.5% but their evaluation method is slightly different. They achieve best result 98.94% when clothing and load variations are included. Chang et al. [107] evaluated their method on real-time videos to achieve 84.38% after using the CASIA dataset for training. Chang and Wu [108] achieved 94% average accuracy for multi-view sequences without requiring prior knowledge of the view angle.

For the IRIP gait dataset, Hu et al. [118] reported state of the art performance of 98.33% using side view sequences only. For multiview sequences, Chen et al. [109] achieved 93.3% from fusion of views, which would require using a camera for each view. This would increase the computing complexity of the system and limit real-world application [136].

In cross database testing, Yu et al. [117] found that the performance decreases. For example, training with the CASIA dataset, only 87.15% accuracy could be achieved with the SOTON dataset, and 87.9% vice versa. Both dataset have different ethnicity of subjects, as well as clothing and capture conditions.

As a conclusion, gait-based gender recognition can achieve high classification rate in controlled datasets, especially with a single view, usually side view. There is a need for more investigation into generalization ability (cross database testing) and performance for datasets containing larger number of subjects with sequences taken under unconstrained environments.

## 5    Gender Recognition by Body

Here, we refer to the use of the static human body (either partially or as a whole) in an image to infer gender, as opposed to using the face region only. Results from a sequence of several images may be fused. Classifying gender based on the human body, like gait, is useful in situations where using the face is not possible or preferred, for example, insufficient resolution or back view [137].

### 5.1    Challenges

Gender recognition based on human body is challenging in several aspects. To infer the gender of a person, humans use not only body shape and hairstyle, but additional cues such as type of clothes and accessories [138]. However, people of the same gender may choose different styles of clothes [139]. On the other hand, the clothing styles worn by both males and females may be somewhat similar. The same is true for hairstyles. The classifier should also be robust to variation, pose, articulation and occlusion of the person. Images of a person may also be taken under different illumination and background clutter. Perhaps due to difficulty caused by such variety, there are few work found in literature on body-based gender recognition.

### 5.2    Feature extraction

The first attempt to recognize gender from full body images was by Cao et al. [139]. The human image is first centered and aligned so the height is normalized as a preprocessing step, and then partitioned into patches corresponding to some parts of the body. Each part was represented using Histogram of Oriented Gradients (HOG) feature, which was previously developed for human detection in images [3].

To compute HOG features, the image gradient is first obtained by using a 1-D mask [-1,0,1] in horizontal and vertical direction. Next, the image is divided into cells and the orientation histogram of gradients is computed for each cell. The cells are grouped into larger overlapping blocks and each block is normalized for contrast. The histograms in a block are concatenated to form a feature vector. The HOG feature is able to capture local shape information from the gradient structure with easily controllable degree of invariance to translations or rotations [3]

Collins et al. [140] proposed their descriptor called PixelHOG (PiHOG) using dense HOG features computed from a custom edge map. In addition, color information was captured using a histogram computed based on the hue and saturation value of the pixels. Their descriptor was found to perform better when compared to those based on Pyramid Histogram of Orientation Gradients [141] and Pyramid Histogram of Words [142].

Bourdev et al. [138] used a set of patches they call *poselet* for inferring attributes of people in unconstrained environments. HOG features, color histogram and skin features were used to represent the poselets. The poselets were used to train attribute classifiers which were combined together to infer gender using context information. Their approach is robust to variations in pose and occlusion, but requires a training dataset with detailed annotations of keypoints of the human body.

Biologically-inspired features (BIF) were used for human body gender recognition by Guo et al. [137]. Only C1 features were used, as it was found that C2 features degraded performance (as in the case of face gender recognition). Various manifold learning techniques was applied on the features. Best results were obtained by first classifying the view (front, back, or mixed) using BIF with PCA, and followed by the gender classifier. PCA was more effective for mixed view, while LSDA (Locally Sensitive Discriminant Analysis) for front and back view.

### 5.3 People Datasets

Table 5 summarizes several publicly available datasets that has been used. The MIT and VIPeR datasets contain images of upright people only. The more challenging Attributes of People dataset contains images taken from the H3D dataset [143] and the PASCAL 2010 *trainval* set for the person category (using the high resolution versions from Flickr where available). The gender is labeled for 5760 individuals but for the rest it remains unspecified.

**Table 5.** Public people datasets

| Dataset | No. of images | No. of unique individuals | Views |
|---|---|---|---|
| MIT CBCL [144] | 924 | < 924 | Front, back |
| VIPeR [145] | 1264 | 632 | Front, back, side, diagonal |
| PASCAL VOC – Person [146] | 4015 | 9218 | Uncontrolled |
| Attributes of People [138] | 8035 | 8035 (5760 gender labeled) | Uncontrolled |

### 5.4 Evaluation and results

Table 6 summarizes the results obtained from works on body-based gender recognition.

**Table 6.** Body-based gender recognition

| First Author, Year | Feature Extraction | Classifier | Training data | Test data | Ave. Acc.% | Dataset variety |
|---|---|---|---|---|---|---|
| Cao, 2008 [139] | HOG | Adaboost variant | MIT-CBCL 600m 288f | 5-CV | 75 | View (frontal, back) |
| Collins, 2009 [140] | PiHOG, colour | SVM-linear | MIT-CBCL 123m 123f | 5-CV | 76 | View (frontal) |
| | | | VIPeR 292m 291f | 5-CV | 80.62 | View (frontal) |
| Guo, 2009 [137] | BIF+PCA/LSDA | SVM-linear | MIT-CBCL 600m 288f | 5-CV | 80.6 | View (frontal, back) |
| Bourdev, 2011 [138] | HOG, colour histogram, skin pixels | SVM | Attributes of People 3395m 2365f (split between training, validation & test sets) | | 82.4 | Unconstrained |

Bourdev et al. [138] achieved 82.4 % accuracy with unconstrained images of people, but the ratio of male and female in their dataset is not balanced. Collins et al. [140] achieved 80.6 % accuracy on a more balanced but small dataset with frontal view only. From these results, there is still room for improvement, although it is not clear how much.

### 5.5 3-D methods

Balan and Black [147] proposed a method to estimate the 3-D shape of a person from images of the person wearing clothing. Multiples poses are required from views of four synchronized cameras and a model of human body shapes learned from a database of range scans is used to infer 3-D body shape. Using a dataset with a small number of subjects, gender was classified using body shape with 94.3% accuracy. In a further work, Guan et al. [148] [149] proposed methods to estimate body shape from a single image. However, the approximate viewing direction and pose are assumed to be known and would be vulnerable to occlusions [148].

Wuhrer and Rioux [150] proposed a posture invariant technique to classify gender based on human body shape using triangular meshes obtained from laser range scanners. Geodesic distances between landmarks on the body were used as features for SVM classifier which achieved at least 93% classification accuracy.

Tang et al. [151] investigated gender recognition using 3-D human body shapes from laser scanning. Mesh normal distribution, curvature information (including mean and Gaussian curvatures), circle-based Fourier descriptors were used as features. When these features were combined and SVM with RBF kernel used as classifier, 98.3% accuracy was achieved on a dataset of 1225 males and 1259 females in standing posture.

## 6 Conclusion

In this paper, we have presented a comprehensive survey on human gender recognition using computer vision-based methods, focusing on 2-D approaches. A lot of work has been done utilizing facial information, with comparatively less exploiting the whole body, whether from motion sequences such as gait or still image. We have also highlighted the challenges and confounding factors, as well as provide a review of the commonly-used features. Face-based gender recognition can be categorized into geometric-based and appearance-based methods, with the latter dominating for the past decade. Good performance has been achieved for images of frontal faces. For multi-view situations (involving both frontal and non-frontal faces), there is room for improvement, especially in uncontrolled conditions, as required in many practical applications. The impact of age on classification rate has been identified in several studies, and more work can be done to improve on this.

In situations when facial analysis is unsuitable, we can turn to using the whole body. Current gait-based gender recognition, whether model-based or appearance-based, depend on the availability of one or more complete gait sequences. High classi-fication rate have been achieved with controlled datasets, especially with side views. Investigation of the generalization ability of the methods (through cross database testing) is called for. Performance for datasets containing larger number of subjects with sequences taken under unconstrained environments is not yet established. To this end, the suitable dataset need to be collected and made available. Some work has also been done based on the static human body (either partially or as a whole) in an image to infer gender.